\documentclass[11pt]{article}

\usepackage[final]{acl}

\usepackage{times}
\usepackage{latexsym}

\usepackage[T1]{fontenc}

\usepackage[utf8]{inputenc}
\usepackage[table]{xcolor} 

\usepackage{microtype}

\usepackage{inconsolata}

\usepackage{graphicx}
\usepackage{float}
\usepackage{amsmath}
\usepackage{amssymb}
\usepackage{booktabs}
\usepackage{multirow} 
\usepackage{bbding}
\usepackage{makecell}
\usepackage{xcolor}
\usepackage{listings}
\usepackage{tabularx}
\usepackage{tcolorbox}
\usepackage{pifont}
\usepackage{hyperref}
\usepackage{balance}

\title{WikiSeeker: Rethinking the Role of Vision-Language Models in Knowledge-Based Visual Question Answering}

\author{
 \textbf{Yingjian Zhu\textsuperscript{1,2}},
   \textbf{Xinming Wang\textsuperscript{1,2}},
    \textbf{Kun Ding\textsuperscript{2}},
  \textbf{Ying Wang\textsuperscript{2}},
\\
 \textbf{Bin Fan\textsuperscript{3}},
  \textbf{Shiming Xiang\textsuperscript{2,\thanks{Corresponding author}}}
\\
 \textsuperscript{1}School of Artificial Intelligence, University of Chinese Academy of Sciences,
\\
 \textsuperscript{2}State Key Laboratory of Multimodal Artificial Intelligence Systems (MAIS),
 \\
 Institute of Automation, Chinese Academy of Sciences,
\\
 \textsuperscript{3}School of Artificial Intelligence, University of
Science and Technology Beijing
\\
 {\normalsize \{zhuyingjian2024, wangxinming2024\}@ia.ac.cn, \{kun.ding, smxiang\}@nlpr.ia.ac.cn}
}

\begin{document}
\maketitle
\begin{abstract}
Multi-modal Retrieval-Augmented Generation (RAG) has emerged as a highly effective paradigm for Knowledge-Based Visual Question Answering (KB-VQA). Despite recent advancements, prevailing methods still primarily depend on images as the retrieval key, and often overlook or misplace the role of Vision-Language Models (VLMs), thereby failing to leverage their potential fully. In this paper, we introduce WikiSeeker, a novel multi-modal RAG framework that bridges these gaps by proposing a multi-modal retriever and redefining the role of VLMs. Rather than serving merely as answer generators, we assign VLMs two specialized agents: a Refiner and an Inspector. The Refiner utilizes the capability of VLMs to rewrite the textual query according to the input image, significantly improving the performance of the multimodal retriever. The Inspector facilitates a decoupled generation strategy by selectively routing reliable retrieved context to another LLM for answer generation, while relying on the VLM's internal knowledge when retrieval is unreliable. Extensive experiments on EVQA, InfoSeek, and M2KR demonstrate that WikiSeeker achieves state-of-the-art performance, with substantial improvements in both retrieval accuracy and answer quality. Our code will be released on \href{https://github.com/zhuyjan/WikiSeeker}{https://github.com/zhuyjan/WikiSeeker.}

\end{abstract}

\section{Introduction}

\begin{figure}[htbp]
\centering
\includegraphics[width=0.9\columnwidth]{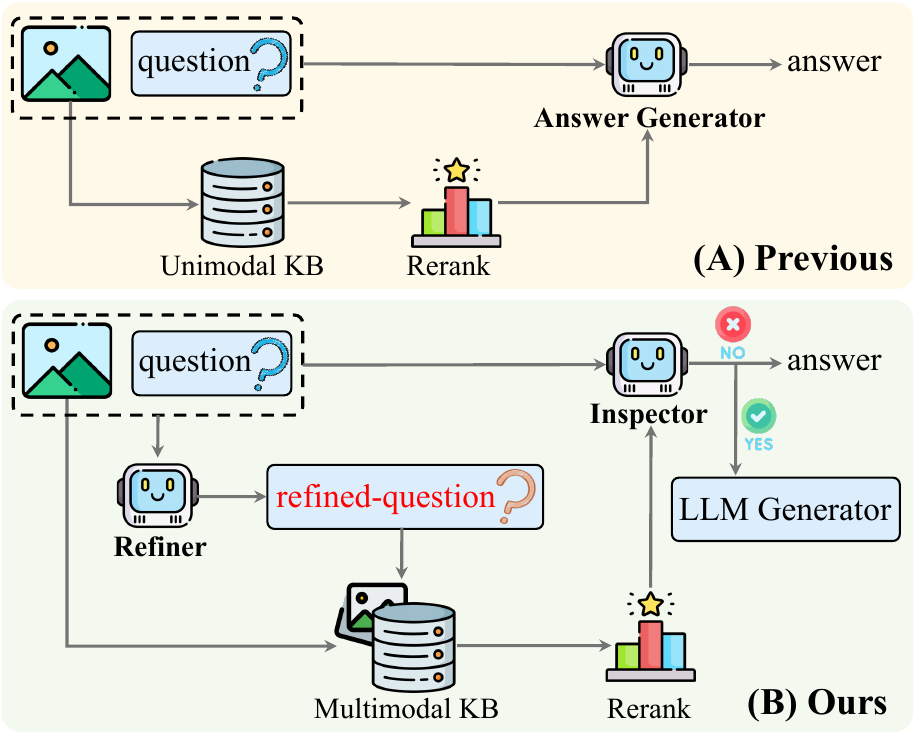}
\caption{The overall architecture of WikiSeeker in comparison with previous methods. (A) Existing methods typically rely on visual-only retrieval and employ the VLM only as the answer generator. (B) Our proposed WikiSeeker conducts multi-modal retrieval and redefines the VLM as a Refiner and an Inspector.}
\label{fig1}
\end{figure}

Retrieval-Augmented Generation (RAG) has been extensively validated as an effective paradigm for addressing Knowledge-Based Visual Question Answering (KB-VQA)~\cite{echosight, omgm, reflectiva}, which necessitates external context or information not present in the image. Such progress is closely intertwined with the broader evolution of autonomous agents~\cite{wang2025hitchhiker} and aligned reasoning models~\cite{wang2025mr}, as well as rapid developments in fine-grained multi-modal visual perception~\cite{zhu2026seavis, chen2025sam}.  As illustrated in Figure~\ref{fig1}(A), existing methods typically utilize the query image as the retrieval key to obtain relevant documents from a knowledge base. These retrieved documents are then concatenated with the input query and passed to an answer generator to produce the final response.  However, existing approaches under this paradigm suffer from the following two critical limitations:

\noindent
\textbf{Visual-Only Retrieval.} Most prior methods rely on a visual-only retrieval strategy~\cite{echosight, omgm, coremmrag}, in which the retrieval key is restricted to the query image. This design overlooks the semantic information contained in the user’s textual query during the retrieval stage, often resulting in irrelevant results when the visual content is ambiguous.

\noindent
\textbf{Misplaced VLM Role.} In multimodal RAG systems, VLMs are typically used solely as answer generators~\cite{wikillava,omgm,echosight}. However, our empirical analysis shows that VLMs are less effective than textual Large Language Models (LLMs) at summarization and extracting correct answers from the retrieved context, as illustrated in Table~\ref{tab:llmbetter}. Consequently, existing methods fail to fully leverage the potential of VLMs within multimodal RAG systems.

To address these limitations, we introduce WikiSeeker, a novel multi-modal RAG framework as illustrated in Figure~\ref{fig1}(B). To tackle the first issue of the visual-only retrieval, we propose a multi-modal knowledge base that allows textual queries to participate in the retrieval process. Furthermore, to resolve the misplaced role of VLMs, instead of using the VLM as a generic answer generator, we reposition it into two specialized agents:

\noindent
\textbf{Refiner}: The VLM leverages visual cues to rewrite the original question, generating a more specific query that explicitly captures visual entities and better aligns with the user's intent.

\noindent
\textbf{Inspector}: After retrieval, the VLM evaluates whether the retrieved context is sufficient to answer the question. If the inspection passes, the refined query is routed to an LLM generator to produce the final answer; otherwise, the VLM answers directly using its internal parametric knowledge.

In summary, our main contributions can be mainly summarized as follows:
\begin{itemize}
    \item We propose a multi-modal knowledge base and introduce a weighted dense retrieval strategy that flexibly integrates visual and textual features for multi-modal retrieval.
    \vspace{-2pt}
    \item We introduce a novel architecture that employs VLMs as specialized agents. Specifically, we design a Reinforcement Learning (RL)-based Refiner that leverages visual cues to rewrite the original question, and an Inspector to validate retrieved context.
    \vspace{-2pt}
    \item Extensive experiments on three widely used benchmarks including EVQA, InfoSeek and M2KR demonstrate that WikiSeeker outperforms existing methods across all main metrics, achieving state-of-the-art performance.
\end{itemize}

\section{Related Work}
\subsection{KB-VQA}
Knowledge-Based VQA (KB-VQA) requires external sense or world knowledge beyond visual recognition. Early benchmarks such as KVQA, FVQA, and KB-VQA~\cite{kvqa,fvqa,kbvqa} were designed to target knowledge-intensive questions, but the required knowledge in these datasets is completely “closed”. Subsequent datasets like OK-VQA~\cite{okvqa} significantly expanded the scale and visual-semantic diversity. Moreover, A-OKVQA~\cite{aokvqa}  models perform reasoning rather than simple fact retrieval to answer questions.

Recent works have introduced more challenging benchmarks that integrate encyclopedic knowledge with extensive multimodal information, specifically E-VQA and InfoSeek~\cite{evqa, infoseek}. Furthermore, M2KR~\cite{preflmr} unifies multiple visual knowledge retrieval tasks into a comprehensive framework. Our work achieves better performance on these challenging benchmarks compared to existing methods.

\subsection{Multi-modal RAG}
Multi-modal Retrieval-Augmented Generation (RAG) is effective to address the limited knowledge coverage and hallucination issues of multimodal large language models (MLLMs) on KB-VQA by incorporating external knowledge bases. Wiki-LLaVA~\cite{wikillava} firstly proposed a hierarchical retrieval pipeline to achieve effective retrieval. EchoSight~\cite{echosight} and OMGM~\cite{omgm} focus on implementing and enhancing multimodal rerankers to improve both retrieval and generation performance. Addressing the challenge of noise, LLM-RA~\cite{llmra} and RoRA-VLM~\cite{roravlm} employ LLM-based and similarity-based methods, respectively, to filter irrelevant information from queries and retrieved results. 

\begin{figure*}[t]
\centering
\includegraphics[width=1\textwidth]{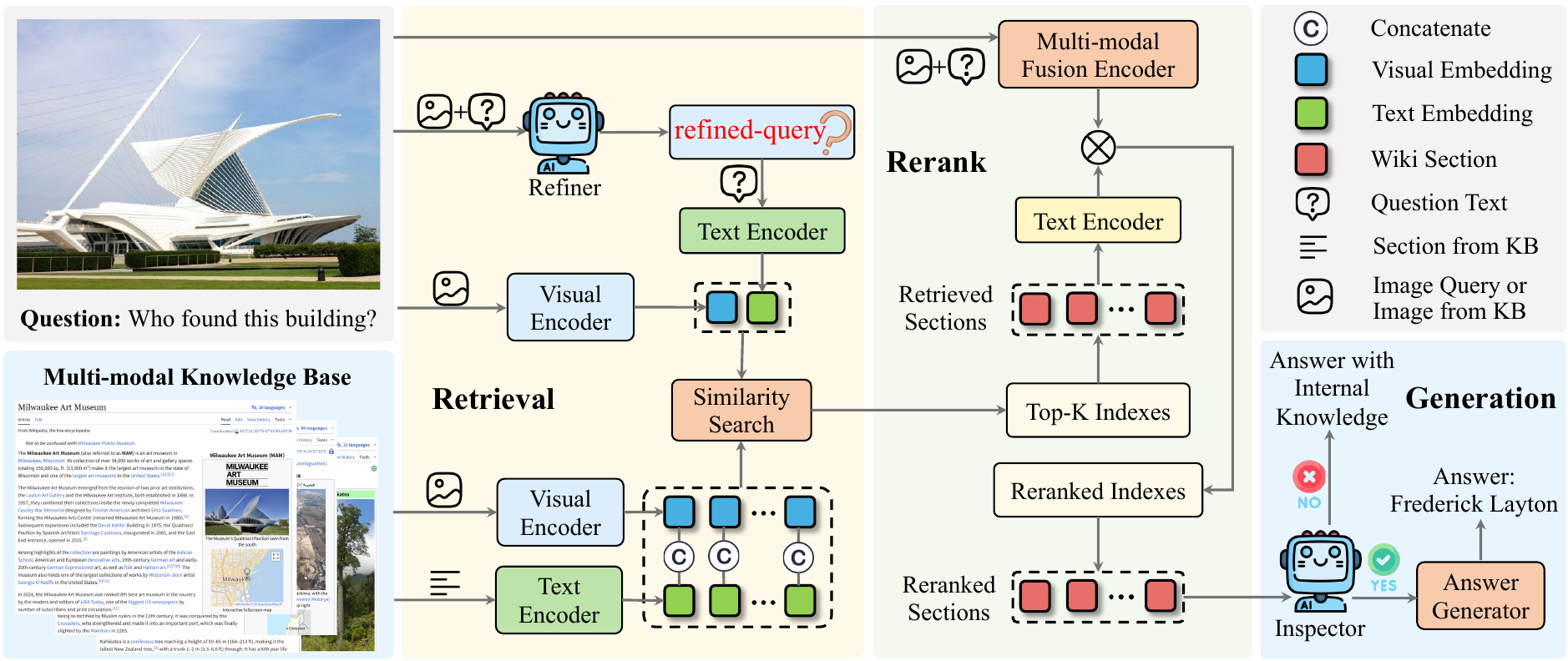}
\caption{The overall architecture of WikiSeeker. We employ VLMs as specialized agents rather than just generators. The pipeline consists of three stages: (1) Retrieval: A VLM-based Refiner expands the initial question with visual semantics. We then perform dense retrieval by concatenating visual and textual embeddings to match against a pre-constructed multi-modal knowledge base. (2) Rerank: A multi-modal reranker filters the top relevant candidate sections. (3) Generation: A VLM-based Inspector evaluates the sufficiency of the retrieved context. It dynamically routes the query: valid contexts are processed by Answer Generator (path ``YES''), while insufficient contexts trigger the Inspector to answer using internal parametric knowledge (path ``NO'').}
\label{fig2}
\end{figure*}

Moving towards more autonomous systems, mR$^2$AG~\cite{mr2ag} and ReflectiVA~\cite{reflectiva} introduce reflection mechanisms into the RAG system, allowing models to dynamically decide whether external knowledge is necessary and identify which retrieved content is valid, thereby enhancing accuracy. To further resolve knowledge source reliability, CoRe-MMRAG~\cite{coremmrag} proposes a cross-source reconciliation framework that addresses inconsistencies between parametric and retrieved knowledge, as well as misalignments between visual and textual modalities, through a four-stage integration pipeline. Most recently, MI-RAG~\cite{mirag} firstly introduced an iterative framework to KB-VQA, enabling the progressive refinement of retrieval and reasoning. Distinct from these approaches, our work focuses on repositioning VLMs within the Multi-modal RAG framework, aiming to fully leverage their potential in both retrieval and generation stages.

\subsection{Reinforcement Learning}
RL methods were introduced to LLM tuning through RL from human feedback (RLHF) via Proximal Policy Optimization~\cite{ppo}. However, PPO requires multiple rounds of LLM optimization, making it challenging to implement. To simplify RL-based tuning, methods such as Direct Preference Optimization (DPO)~\cite{dpo} and SimPO~\cite{simpo} have been proposed. Group Relative Policy Optimization (GRPO)~\cite{grpo} removes the need for a critic model entirely by utilizing the average reward of a group of outputs as the baseline, which is utilized to train our Refiner.

\section{Methodology}
To address the limitations of current multi-modal RAG systems, we propose WikiSeeker as illustrated in Figure~\ref{fig2}. WikiSeeker performs multi-modal retrieval and redefines the VLM as a Refiner and an Inspector. The Refiner leverages the VLM's capabilities to rewrite and expand the original question based on the query image. Following refinement, the system performs dense retrieval using a multi-modal embedding strategy. Specifically, the refined text query and input image are encoded separately and then concatenated to form a unified query embedding. This embedding is matched against an index of our knowledge base, which consists of similarly concatenated embeddings of knowledge base (KB) images and section texts. After retrieving the top candidates, we employ a multi-modal reranker to filter for the most relevant sections. Finally, the Inspector evaluates whether the retrieved information is sufficient and consistent.  If the inspection passes, the answer generator produces the final response; otherwise, it answers directly using its internal knowledge.

\subsection{Multi-modal Dense Retrieval}
\noindent
\textbf{Multi-modal Index Construction.} Unlike previous approaches~\cite{echosight, omgm} that construct indices based on isolated images or coarse-grained article-level summaries, we construct a fine-grained Knowledge Base (KB) composed of aligned <$\text{image}, \text{section}$> pairs. For every image $I_{kb}$ in the source data, we identify its corresponding textual section $T_{kb}$. To ensure semantic density and fix length variations, we employ an LLM to summarize excessively long sections. For images that correspond to missing content or low-information sections (e.g., reference lists), we substitute $T_{kb}$ with the abstract section of the article. We then utilize a visual encoder $\Phi_{\text{vis}}$ and a textual encoder $\Phi_{\text{text}}$ to generate the index vector $\mathbf{v}_i$ by directly concatenating the visual and textual embeddings of the $i$-th entry:
\begin{equation}
\label{eq:0}
\mathbf{v}_i=\text{Concat}\left[\Phi_{\text{vis}}(I_{kb}), \Phi_{\text{text}}(T_{kb})\right].
\end{equation}

\noindent
\textbf{Embedding and Retrieval.} To fuse the modalities and flexibly control the semantic contribution of each modality, we employ a weighted concatenation strategy. We introduce a hyperparameter $\alpha \in [0, 1]$ to control the relative importance of visual and textual features. During the retrieval phase, given an input query image $I_q$ and the refined question $T_q$, the weighted multi-modal embedding vector $\mathbf{v}_q$ is defined as:
\begin{equation}
\label{eq:1}
\mathbf{v}_q=\text{Concat}\left[\alpha\cdot \Phi_{\text{vis}}(I_q), (1-\alpha)\cdot\Phi_{\text{text}}(T_q)\right].
\end{equation}
We then compute the cosine similarity between the query and all entries in the knowledge base to retrieve the top-$k$ most relevant sections. The retrieval set $\mathcal{S}_{\text{ret}}$ is obtained as follows:
\begin{equation}
\label{eq:2}
\mathcal{S}_{\text{ret}}\!=\!\left\{s_i\!=\!\left\langle\! \frac{\mathbf{v}_q\!\cdot\!\mathbf{v}_i}{||\mathbf{v}_q||\!\cdot\!||\mathbf{v}_i||}\!\right\rangle, i = 1, \dots, N \right\}_{\text{top-}k},
\end{equation}
where $\mathbf{v}_i$ represents the multi-modal vector of the $i$-th entry in the knowledge base.

In Knowledge-Based Visual Question Answering (KB-VQA), user queries are characteristically concise and abstract. This inherent ambiguity often introduces excessive noise during the multimodal retrieval process, thereby significantly degrading retrieval quality. To mitigate this, we propose a VLM Refiner that leverages visual cues to rewrite and expand user queries. Unlike traditional approaches that rely on supervised fine-tuning with expensive human-annotated query pairs, our framework adopts a Reinforcement Learning (RL) paradigm. This allows the model to autonomously discover optimal query refinement strategies by interacting with the retrieval system and receiving feedback based on retrieval performance.

\subsection{VLM as Refiner}
\begin{figure}[htbp]
\centering
\includegraphics[width=1.0\columnwidth]{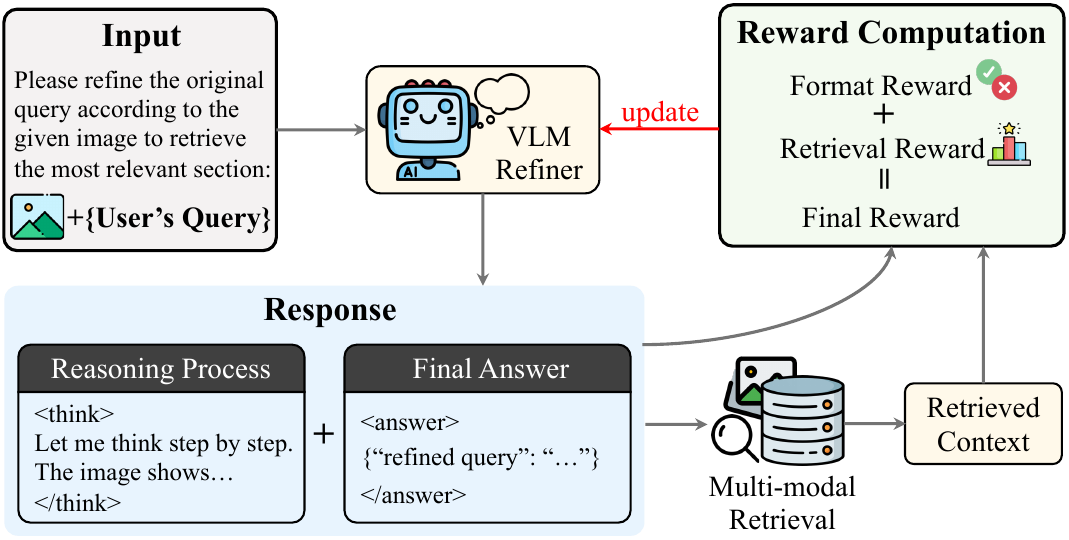}
\caption{The training pipeline of the VLM Refiner via Reinforcement Learning. The refined query is used to retrieve information from the multimodal knowledge base. VLM’s response is used to compute the structure-based format reward, while the retrieved context is used to compute the rank-based retrieval reward. The combined reward is finally used to update the model through the GRPO algorithm.}
\label{fig3}
\end{figure}

\noindent
\textbf{Reasoning-Enhanced Generation} To promote a deeper understanding of the visual content, we enforce a structured generation process. Instead of directly producing a refined query, the model is encouraged to first generate a chain-of-thought (CoT)~\cite{cot} reasoning sequence enclosed in <think> tags, followed by the final output in <answer> tags. This mechanism enhances the robustness and interpretability of the query refinement process through explicit visual reasoning.

\noindent
\textbf{Optimization via GRPO} We optimize the VLM policy $\pi_\theta$ using Group Relative Policy Optimization (GRPO)~\cite{grpo}. For each query $q$, GRPO samples a group of $G$ outputs $\{o_1, \dots, o_G\}$ from the old policy $\pi_{\theta_{\text{old}}}$. The optimization objective can be defined as:
\begin{equation}
\label{eq:3}
\begin{aligned}
& \mathcal{J}_{\text{GRPO}}(\theta)
= \mathbb{E}_{q \sim \mathcal{D},\, \{o_i\}_{i=1}^G \sim \pi_{\theta_{\text{old}}}}
  \Bigg[
    \frac{1}{G} \sum_{i=1}^{G}
    \Big(
      \min\bigl( \rho_i A_i,\\
                 &\operatorname{clip}(\rho_i, 1-\epsilon, 1+\epsilon)\, A_i
          \bigr)
- \beta \,\mathbb{D}_{\text{KL}}\bigl(\pi_\theta \,\|\, \pi_{\text{ref}}\bigr)\Big)
  \Bigg],
\end{aligned}
\end{equation}
where $\epsilon$ and $\beta$ are hyper-parameters, and $\rho_i = \frac{\pi_\theta(o_i | I, q)}{\pi_{\theta_{\text{old}}}(o_i | I, q)}$ is the probability ratio. Crucially, $A_i$ is the advantage, computed using the group of rewards $\{r_1, r_2, \dots, r_G\}$ corresponding to the outputs within each group:
\begin{equation}
\label{eq:4}
A_i = \frac{r_i - \text{mean}(\{r_1, r_2, \dots, r_G\})}{\text{std}(\{r_1, r_2, \dots, r_G\})}.
\end{equation}

\noindent
\textbf{Reward Function Design}
Inspired by DeepRetrieval~\cite{deepretrieval}, the reward signal $r_i$ for the $i$-th generated output $o_i$, is composed of two distinct components: retrieval reward $r_{\text{retrieval}}(o_i)$ and format reward $r_{\text{format}}(o_i)$. Formally, the total reward is defined as:
\begin{equation}
\label{eq:5}
r_i = r_{\text{retrieval}}(o_i) + r_{\text{format}}(o_i).
\end{equation}
The format reward measures the syntactic correctness of the generated text. We check both the presence and the correct ordering of the XML-style tags. In addition, the content inside <answer> must be a valid and parseable JSON object that includes the required key. A positive score of +1 is assigned when the output fully satisfies these constraints, whereas any structural violation results in a substantial penalty of -4.
\begin{table}[htbp]
    \centering
    \small
    \setlength{\tabcolsep}{2pt}
    \begin{tabular}{lcccccc}
        \toprule
        hit rank & $\left[1,5\right]$ & $\left[6,10\right]$ & $\left[11,20\right]$ & $\left[21,50\right]$ & $\left[51,100\right]$ & $\left[101,200\right]$ \\
        \midrule
        reward   & +4        & 
        +3.5       & +3         & +1         & +0.5        & +0.1  \\
        \bottomrule
    \end{tabular}
    \caption{The discrete mapping of retrieval rewards based on the hit rank.}
    \label{method:tab1}
\end{table}
For the retrieval reward, we perform multimodal retrieval using the refined query generated by the VLM together with the image query. We obtain the top 200 retrieved Wikipedia entities, and if the ground-truth entity is hit, a reward is assigned based on the criteria defined in Table~\ref{method:tab1}. Otherwise, a penalty of -2.5 is applied.

\subsection{VLM as Inspector}
\begin{table}[hbp]
\centering
\small
\setlength{\tabcolsep}{3.0pt} 
\begin{tabular}{lcccc}
\toprule
\textbf{Method} & Ratio=0 & Ratio=0.3 & Ratio=0.7 & Ratio=1.0 \\
\midrule
QwenVL(I+T) & \textbf{22.75} & 41.26 & 68.99 & 88.46 \\
QwenVL(T) & 19.44 & 41.75 & 70.82 & 91.10 \\
LLaMA & 18.51 & 40.48 & \textbf{70.88} & 92.99 \\
Qwen & 19.77 & \textbf{42.52} & 70.80 & \textbf{93.45} \\
\bottomrule
\end{tabular}
\caption{VQA performance comparison under different signal-to-noise ratios. The ratio indicates the proportion of correct sections in the input context. QwenVL(I+T) and QwenVL(T) denote Qwen2.5-VL with and without image input, respectively. More detailed experimental results can be found in Table~\ref{supp:tab:snr_full} in Appendix~\ref{supp:limitVLM}.}
\label{tab:llmbetter}
\end{table}

Existing KB-VQA methods~\cite{omgm, wikillava} typically use VLMs as answer generators. However, our empirical analysis shows that VLMs are less effective than LLMs at leveraging retrieved context to answer questions. As illustrated in Table~\ref{tab:llmbetter}, we compare Qwen2.5-VL-7B (with and without image input) against two strong text-only baselines: LLaMA3.1-8B~\cite{llama} and Qwen2.5-7B. As shown in the table, as retrieval performance gradually improves, starting from 0.3, the performance of QwenVL with image input consistently falls below that of all text-only models. This suggests that in KB-VQA task where the answer is contained in text evidence, visual tokens from the query image often behave as noise rather than helpful context and may distract the model from leveraging the retrieved information.

Based on this observation, we propose a decoupled generation strategy that explicitly separates visual perception from textual reading comprehension. In our framework, an LLM is responsible for precise answer extraction, while the VLM functions as an inspector to validate the retrieved context and provide the answer using its internal parametric knowledge if the inspection is not passed.

The Inspector ($\mathcal{M}_{\text{ins}}$) takes the query image $I_q$, the question $Q$, and the reranked sections $\mathcal{S}_{\text{rerank}}$ as input. Its goal is to determine whether the retrieved context is sufficient to answer the question and consistent with the visual evidence. Formally, the inspection process can be formulated as:
\begin{equation}
\label{eq:6}
(s, A_{\text{internal}}) = \mathcal{M}_{\text{ins}}(I_q, Q, \mathcal{S}_{\text{rerank}}),
\end{equation}
where $s \in \{ \text{PASS}, \text{FAIL} \}$ denotes the decision result, indicating whether the inspection is passed, and $A_{\text{internal}}$ represents the answer generated directly by the VLM using its parametric knowledge.

If the context passes inspection ($s = \text{PASS}$), we rely on the retrieved context and use a text-only LLM ($\mathcal{M}_{\text{gen}}$) to generate the final answer $A$. If inspection fails ($s = \text{FAIL}$), we instead select the VLM’s answer. This process can be expressed as:
\begin{equation}
\label{eq:7}
A = \begin{cases} \mathcal{M}_{\text{gen}}(Q, \mathcal{S}_{\text{rerank}}), & \text{if } s = \text{PASS} \\ A_{\text{internal}}, & \text{if } s = \text{FAIL} \end{cases}.
\end{equation}

\begin{table*}[t]
\fontsize{9}{10}\selectfont
  \centering
  \setlength{\tabcolsep}{3mm}
  \resizebox{\textwidth}{!}{
  \begin{tabular}{l cccc c cccc}
   \toprule
    \multirow{2}{*}[-0.6ex]{\textbf{Method}}& \multicolumn{4}{c}{\textbf{E-VQA}} & & \multicolumn{4}{c}{\textbf{InfoSeek}} \\
    \cmidrule{2-5} \cmidrule{7-10} 
     & R@1 & R@5 & R@10 & R@20 & & R@1 & R@5 & R@10 & R@20 \\
    \midrule
    CLIP I-T & 3.3 & 7.7 & 12.1 & 16.5 & & 32.0 & 54.0 & 61.6 & 68.2 \\
    Wiki-LLaVA & 3.3 & - & 9.9 & 13.2 & & 36.9 & - & 66.1 & 71.9 \\
    LLM-RA & - & - & - & - & & 47.3 & 53.8 & - & - \\
    mR$^2$AG & - & - & - & - & & 38.0 & - & 65.0 & 71.0 \\
    ReflectiVA & 15.6 & 36.1 & - & 49.8 & & 56.1 & 77.6 & - & 86.4 \\
    EchoSight & 36.5 & 47.9 & 48.8 & 48.8 & & 53.2 & 74.0 & 77.4 & 77.9 \\ %
    CoRe-MMRAG & 13.3 & 31.3 & 41.0 & - & & 45.6 & 67.1 & 73.0 & - \\ %
    OMGM & 42.8 & 55.7 & 58.1 & 58.7 & & 64.0 & 80.8 & 83.6 & 84.8 \\
    WikiSeeker (ours) \\
    \hspace{0.5em}\textit{w/o. Refiner} & 28.0 & 37.2 & 40.9 & 43.4 & & 53.5 & 74.6 & 77.8 & 78.5 \\
    \rowcolor[HTML]{E6F2FF}
    \hspace{0.5em}\textit{w. Refiner} & \textbf{44.1} & \textbf{59.9} & \textbf{62.1} & \textbf{62.3} & & \textbf{67.0} & \textbf{83.7} & \textbf{86.9} & \textbf{87.7} \\ %
  \bottomrule
  \end{tabular}}
  \caption{Retrieval results on the E-VQA test set and InfoSeek validation set. ``w/o. Refiner'' and ``w. Refiner'' indicate whether the input query is expanded by Refiner. Best results are highlighted in bold.}
  \label{tab:retrieval_main_results_1}
\end{table*}

\begin{table*}[htbp]
\centering
\setlength{\tabcolsep}{1.8mm} 
\fontsize{9}{10}\selectfont
\resizebox{\textwidth}{!}{
\begin{tabular}{l ccccc ccccc cccc}
\toprule
\multirow{2}{*}[-0.6ex]{\textbf{Method}} &
\multicolumn{5}{c}{\textbf{E-VQA (M2KR)}} &
\multicolumn{5}{c}{\textbf{OKVQA-GS (M2KR)}} &
\multicolumn{4}{c}{\textbf{OVEN (M2KR)}} \\
\cmidrule(lr){2-6} \cmidrule(lr){7-11} \cmidrule(lr){12-15}
 & R@1 & R@5 & R@10 & PR@5 & PR@20
 & R@1 & R@5 & R@10 & PR@5 & PR@20
 & R@1 & R@5 & R@10 & R@20 \\
\midrule
CLIP & 3.3 & 7.7 & 12.1 & 10.4 & - & - & - & - & 5.7 & - & - & 22.0 & - & - \\
FLMR & - & - & - & - & - & - & - & - & 68.1 & - & - & 40.5 & - & - \\
PreFLMR \\
\textit{w/o. Refiner}& 40.4 & 62.5 & 70.2 & 72.2 & 78.0 & 13.8 & 30.5 & 39.3 & 67.5 & 86.6 & 31.1 & 64.2 & 76.5 & 85.2 \\
\rowcolor[HTML]{E6F2FF}
\textit{w. Refiner} & \textbf{43.1} & \textbf{65.3} & \textbf{73.2} & \textbf{72.5} & \textbf{79.9} & \textbf{20.7} & \textbf{41.5} & \textbf{51.1} & \textbf{77.5} & \textbf{91.6} & \textbf{42.8} & \textbf{69.7} & \textbf{78.8} & \textbf{86.9} \\
\bottomrule
\end{tabular}
}
\caption{Retrieval performance on M2KR benchmark. ``w/o. Refiner'' and ``w. Refiner'' indicate whether the input query is expanded by Refiner. Best results are highlighted in bold.}
\label{tab:retrieval_main_results_2}
\end{table*}

\section{Experiments}
\subsection{Datasets and Metrics}
We conduct our experiments on three widely used KB-VQA benchmarks: Encyclopedic VQA (EVQA), InfoSeek and M2KR. The detailed dataset information and setup can be found in Appendix. Performance evaluation is conducted along two dimensions. The retrieval performance is measured using Recall@K and Pseudo Recall@K. For question-answering performance, each dataset adopts its own official evaluation metric. Specifically, EVQA is evaluated using the BEM score~\cite{bem}, while InfoSeek uses both the standard and relaxed VQA Accuracy~\cite{vqa, methani2020plotqa}.

\subsection{Implementation Details}


We evaluate WikiSeeker against several state-of-the-art multi-modal RAG frameworks for KB-VQA, including Wiki-LLaVA~\cite{omgm}, EchoSight~\cite{echosight}, ReflectiVA\cite{reflectiva}, and OMGM~\cite{omgm}, among others. We directly use the reranker from EchoSight and keep its weights frozen throughout the experiments. After the reranking stage, only the top-1 section is selected. We then apply bge-reranker-v2-m3~\cite{bge} to perform text reranking over the entire article corresponding to that section, and the resulting top-1 section is used as the context for generation. All fine-tuning processes were conducted using the LlamaFactory framework~\cite{zheng2024llamafactory} to enable efficient LLM fine-tuning. All experiments are conducted on 4 NVIDIA A800 40GB SXM4 GPUs. 

\noindent
\textbf{Retriever} We utilize EVA-CLIP-8B~\cite{evaclip} to encode visual inputs (both query image and KB reference images) and Qwen3-Embedding-0.6B~\cite{qwen3embed} to encode textual inputs (refined question and KB sections). To generate the unified representation for both the input query and the <image, section> entries in the knowledge base, we extract the pooled features from the final layer of each encoder and concatenate them. For efficient large-scale indexing and retrieval, we leverage the FAISS library~\cite{faiss}, employing cosine similarity as the distance metric to identify the top-ranked candidates.

\noindent
\textbf{Refiner} 
We implement the Refiner using Qwen2.5-VL-3B-Instruct~\cite{qwen2.5vl}, optimized via the GRPO algorithm within the HybridFlow (Verl) framework~\cite{verl}. To construct RL training datasets for EVQA and InfoSeek, we first perform retrieval on the respective training sets using our multi-modal retriever. We then sample training queries according to the rank of the ground-truth entity (hit rank). The specific sampling distribution is detailed in Table~\ref{exp:tab1}, resulting in a dataset of 7,000 samples for each benchmark.
\begin{table}[htbp]
    \small
    \centering
    \setlength{\tabcolsep}{2pt}
    \begin{tabular}{lccccc}
        \toprule
        hit rank & $\left[1,5\right]$ & $\left[6,10\right]$ & $\left[11,20\right]$ & $\left[21,200\right]$ & miss \\
        \midrule
        EVQA   & 500        & 
        1000       & 1000         & 2500         & 2000  \\
        InfoSeek   & 0        & 
        500       & 1000         & 2500         & 3000  \\
        \bottomrule
    \end{tabular}
    \caption{The sampling distribution based on retrieval hit rank used to construct the RL datasets.}
    \label{exp:tab1}
\end{table}

\begin{table*}[htbp]
\fontsize{8}{10}\selectfont
\renewcommand{\arraystretch}{1.1}
\centering
\setlength{\tabcolsep}{1mm}
\resizebox{\textwidth}{!}{
\begin{tabular}{l|c|c|c | c | ccc}
\toprule
\multirow{2}{*}{\textbf{Method}} & \multirow{2}{*}{\textbf{Generator Model}} & \multirow{2}{*}{\textbf{Gen. FT}} & \multirow{2}{*}{\textbf{Ret. FT}} & \multirow{2}{*}{\textbf{E-VQA}} & \multicolumn{3}{c}{\textbf{InfoSeek}} \\
 & & & & & \textbf{Unseen-Q} & \textbf{Unseen-E} & \textbf{Overall} \\
\midrule
RoRA-VLM & LLaVA-1.5-7B & \Checkmark & \XSolidBrush & 20.29 & 27.34 & 25.10 & 26.9 \\ 
Wiki-LLaVA & LLaVA-1.5-7B & \Checkmark & \XSolidBrush & 21.8 & 30.1 & 27.8 & 28.9 \\
LLM-RA & BLIP2-Flan-T5XL & \Checkmark & \Checkmark & - & 26.12 & 20.90 & 23.14 \\
EchoSight & Mistral-7B | LLaMA3-8B & \XSolidBrush & \Checkmark & 41.8 & - & - & 31.3\\
mR$^2$AG & LLaVA-1.5-7B & \Checkmark & \Checkmark & - & 40.6 & 39.8 & 40.2\\
ReflectiVA & LLaVA-MORE-8B & \Checkmark & \Checkmark & 35.5 & 40.4 & 39.8 & 40.1\\
OMGM & LLaVA-1.5-7B & \Checkmark & \Checkmark & 50.17 & 43.46 & 43.53 & 43.49\\

\midrule
\multirow{3}{*}{WikiSeeker (ours)} & Qwen3-VL-8B-Instruct  & \XSolidBrush & \Checkmark & 51.81 & 38.7 & 37.69 & 38.19 \\
& Qwen2.5-7B-Instruct & \XSolidBrush & \Checkmark & 52.97 & 35.99 & 34.93 & 35.46\\ 
& Inspector+Qwen2.5-7B-Instruct & \Checkmark & \Checkmark & \textbf{55.62} & \textbf{43.82} & \textbf{45.64}  & \textbf{44.72}\\ 

\bottomrule
\end{tabular}}
\caption{VQA accuracy comparison with the baselines. Gen. FT and Ret. FT indicate whether the generator and retriever of the method were fine-tuned, respectively. Best results are highlighted in bold.} 
\label{tab:vqa_main_results}
\end{table*}

For the EVQA split in M2KR benchmark, we directly deploy the Refiner trained on the standard EVQA dataset without further tuning. However, for the OKVQA and OVEN~\cite{oven} splits, we utilize the PreFLMR retriever and train on the corresponding training splits provided by M2KR. Regarding hyperparameters, we set the global batch size to 32 and the learning rate to $1 \times 10^{-6}$. The vision tower of the VLM remains frozen during the training process. For the GRPO configuration, we set the group size to 5 and the rollout temperature to 0.7, training the model for a total of 600 steps.

\noindent
\textbf{Answer Generator} We choose Qwen2.5-7B-Instruct~\cite{qwen2.5} as the LLM Generator and fine-tune separate models tailored to each dataset. For InfoSeek, we construct a training set of 13,640 samples using the original ground-truth answers as supervision targets. For EVQA, we employ an auxiliary LLM to enrich the ground-truth answers with more descriptive responses, resulting in a dataset of 10,000 samples.

\noindent
\textbf{Inspector} We initialize the Inspector with Qwen3-VL-8B-Instruct~\cite{qwen3vl} and fine-tune it on a mixed dataset of 38,000 samples derived from the training splits of both datasets. The model is fine-tuned to generate structured JSON outputs.  If the retrieved context matches the gold section, the sample is labeled \{"pass": "true"\}. Otherwise, it is labeled \{"pass": "false"\}, and the Inspector is supervised to generate the ground-truth answer using its internal parametric knowledge. More details can be found in Appendix~\ref{supp:inspector} and Figure~\ref{fig: inspector_prompt}.

\subsection{Main Results}
\noindent
\textbf{Retrieval Results.} Table~\ref{tab:retrieval_main_results_1} presents a comparative analysis of retrieval performance on the EVQA and InfoSeek datasets. The results demonstrate that using refined queries for multimodal retrieval provides a substantial performance improvement over using the original queries, establishing the state-of-the-art (SOTA) results across all evaluated metrics. To further validate the generalizability of our approach, we report results on the M2KR benchmark in Table~\ref{tab:retrieval_main_results_2}. Building upon the strong PreFLMR~\cite{preflmr} baseline, integrating our Refiner consistently improves retrieval accuracy on the EVQA, OKVQA-GS, and OVEN splits. Notably, the enriched queries enable PreFLMR to achieve SOTA performance on all splits across all metrics, with Recall@1 exhibiting impressive gains of 6.9 percentage points on OKVQA-GS and 11.7 percentage points on OVEN, respectively.

\noindent
\textbf{VQA Results.} Table~\ref{tab:vqa_main_results} presents a comprehensive comparison of VQA accuracy between WikiSeeker and existing state-of-the-art methods. Under the zero-shot setting, our approach already surpasses the previous SOTA on the EVQA dataset. The performance is further improved when combining the proposed Inspector with a fine-tuned Qwen2.5-7B-Instruct generator. Overall, our method achieves the best results on both datasets, with particularly notable improvements on EVQA, where it exceeds the previous best method by a substantial margin of 5.45 percentage points. We attribute these significant gains to our two core contributions: the Refiner, which substantially enhances retrieval quality, and the Inspector, which equips the generator with visual understanding while preserving the LLM’s superior information-extraction capabilities.

\subsection{Ablation Study}
To comprehensively evaluate WikiSeeker and the contribution of each component, we carry out extensive experiments focusing on both retrieval and overall question-answering performance.

\noindent
\textbf{Component analysis of WikiSeeker.} To isolate the impact of our core modules, we conduct an ablation study as summarized in Table~\ref{abl:tab1}. We first establish a baseline model that performs multimodal retrieval using the original, unrefined query and employs a fine-tuned Qwen2.5-7B-Instruct as the answer generator without inspection. It can be found that integrating the Refiner leads to substantial improvements, boosting accuracy by 11.57\% on EVQA and 11.79\% on InfoSeek. This significant improvement indicates that the enhanced retrieval quality provided by the Refiner is the primary factor driving final KB-VQA performance. Moreover, introducing the Inspector consistently yields additional gains, regardless of whether the input query is refined. These consistent improvements further validate the effectiveness of our decoupled generation strategy. 
\begin{table}[ht]
\small
\setlength{\tabcolsep}{2.5mm}
\fontsize{9}{10}\selectfont
\centering
\begin{tabular}{c c | c c}
\toprule
Refiner & Inspector & E-VQA & InfoSeek \\
\midrule
\XSolidBrush & \XSolidBrush & 42.35 & 31.68 \\
\XSolidBrush & \Checkmark & 46.28 & 33.29 \\
\Checkmark & \XSolidBrush & 53.92 & 43.47 \\ 
\Checkmark & \Checkmark & \textbf{55.62} & \textbf{44.72} \\
\bottomrule
\end{tabular}
\caption{The ablation study on the impact of Refiner and Inspector on the VQA results.}
\label{abl:tab1}
\end{table}

\noindent
\textbf{Effect of Refiner.}
We conduct a comparative analysis to validate that the performance gains of the Refiner in WikiSeeker. As detailed in Table~\ref{abl:tab2}, we compare our refinement method against several strong baselines. We employ Qwen2.5-VL-7B-Instruct to generate both the Image Caption and Zs Expansion variants, prompt can be refered to Appendix~\ref{supp:prompt}. The results indicate that zero-shot expansion provides a moderate improvement over the original query, demonstrating the importance of refining textual queries in KB-VQA tasks. Notably, our WikiSeeker Refiner, although built on a smaller 3B model, achieves substantial performance gains across all metrics and thereby highlights the effectiveness of our reinforcement learning strategy.

\begin{table}[htbp]
    \small
    \centering
    \setlength{\tabcolsep}{3.5pt}
    \begin{tabular}{lcccc}
        \toprule
        Method & R@1 & R@5 & R@10 & R@20\\
        \midrule
        Original Query & 14.59 & 29.43 & 37.14 & 43.35 \\
        Image Caption & 16.11 & 31.58 & 37.81 & 42.93 \\
        Zs Expansion & 18.78 & 34.88 & 42.4 & 47.79 \\
        Zs Expansion+Caption & 18.34 & 34.88 & 42.15 & 47.81 \\
        WikiSeeker Refiner & \textbf{25.45} & \textbf{44.48} & \textbf{52.42} & \textbf{57.98} \\
        \bottomrule
    \end{tabular}
    \caption{Impact of different query formulation strategies on EVQA retrieval performance.}
    \label{abl:tab2}
\end{table}

\noindent
\textbf{Analysis of Decoupled Generation Strategy.} In this analysis, we utilize VLM to denote Qwen3-VL-8B-Instruct~\cite{qwen3vl} and LLM to denote Qwen2.5-7B-Instruct~\cite{qwen2.5}. To evaluate the effectiveness of the decoupled generation strategy, we first conduct an oracle analysis on the EVQA dataset, as shown in Table~\ref{abl:tab3}. In this table, “Decoupled” refers to the strategy in which successfully retrieved entries are explicitly routed to the LLM, while retrieval failures are routed to the VLM. We observe that this hybrid design yields substantial performance gains, regardless of whether the generators are fine-tuned. Building on this observation, Table~\ref{abl:tab4} presents the comparative results of our proposed Inspector. ``Oracle Ensemble'' is defined as the union of correct predictions from both the VLM and LLM, representing the theoretical upper bound of the decoupled strategy. As shown in the table, LLM generation with Inspector achieves 55.62\% accuracy on EVQA and 44.72\% on InfoSeek, surpassing the strongest fine-tuned baselines and clearly demonstrating the effectiveness of the Inspector within WikiSeeker.

\begin{table}[htbp]
\fontsize{9}{10}\selectfont
\renewcommand{\arraystretch}{1.1}
\centering
\setlength{\tabcolsep}{1mm}
\begin{tabular}{lccc}
\toprule
Setting & Generator & Strategy & Accuracy(\%) \\
\midrule
\multirow{3}{*}{Zero-shot} & VLM  & VLM-only & 51.81 \\
& LLM & LLM-only & 52.97 \\ 
& VLM + LLM & Decoupled & \textbf{54.48} \\ 

\midrule
\multirow{3}{*}{Fine-tuned} & VLM  & VLM-only & 52.59 \\
& LLM & LLM-only & 53.92 \\ 
& VLM + LLM & Decoupled & \textbf{56.34} \\ 

\bottomrule
\end{tabular}
\caption{Oracle analysis of the decoupled generation strategy on the EVQA dataset.} 
\label{abl:tab3}
\end{table}

\begin{table}[htbp]
    \small
    \centering
    \begin{tabular}{l|c|cc}
        \toprule
        Method & FT & EVQA & InfoSeek\\
        \midrule
        VLM & \XSolidBrush & 51.81 & 38.19 \\
        LLM & \XSolidBrush & 52.97 & 35.46 \\
        VLM & \Checkmark & 52.59 & 44.06 \\
        LLM & \Checkmark & 53.92 & 43.47 \\
        LLM + Inspector & \Checkmark & \textbf{55.62} & \textbf{44.72} \\
        \midrule
        Oracle Ensemble & \Checkmark & 58.87 & 47.28 \\
        \bottomrule
    \end{tabular}
    \caption{Ablation study on the effectiveness of the WikiSeeker Inspector on EVQA and InfoSeek.}
    \label{abl:tab4}
\end{table}

\section{Conclusion}
In this work, we have proposed the WikiSeeker, a multi-modal RAG framework for KB-VQA, which implements a multi-modal retriever and redefines the role of vision-language models (VLMs). Specifically, we assign VLMs two specialized agents: a reinforcement learning–optimized Refiner for query expansion and an Inspector for retrieved context validation. The refined query produced by the Refiner is utilized alongside the input image to facilitate robust multi-modal retrieval. Furthermore, the Inspector plays a central role in our decoupled generation strategy by effectively integrating the context summarization capabilities of large language models (LLMs) with the visual reasoning strengths of VLMs. Extensive experiments on the EVQA, InfoSeek, and M2KR benchmarks confirm that WikiSeeker achieves state-of-the-art performance, demonstrating significant improvements in both retrieval accuracy and answer generation. These findings provide valuable insights for designing effective multi-modal retrieval systems in future research on KB-VQA tasks.

\section*{Limitations}
While WikiSeeker achieves strong performance, several directions remain for further improvement. First, our current decoupled generation strategy adopts a hard routing rule: instances with successful retrieval are handled by the LLM, whereas retrieval failures are handled by the VLM. Our experiments suggest that this routing mechanism is not necessarily optimal. More effective collaboration between LLM and VLM under a decoupled design therefore warrants further investigation. Second, WikiSeeker currently supports only single-pass retrieval and cannot address multi-hop questions. Incorporating iterative retrieval mechanisms and training multi-step reasoning and answering via reinforcement learning constitutes a promising direction for future work.

\section*{Ethical Statement}
The datasets Encyclopedic VQA (EVQA), InfoSeek, and M2KR , and the models (including the Qwen2.5-VL, Qwen3-VL, and Qwen2.5-Instruct series)  employed in this study are all open-source, thereby incurring no risks associated with licensing. Furthermore, as our research is centered on the Knowledge-Based Visual Question Answering (KB-VQA) domain, it does not entail risks pertaining to human ethics and values.

\section*{Acknowledgments}

This work was supported by the National Natural Science Foundations of China (Grant No.62306310).

\balance
\bibliography{main}

\appendix
\definecolor{lightgray}{gray}{0.95}
\definecolor{deepblue}{RGB}{70,130,180}
\definecolor{deepgray}{RGB}{119,136,153}
\lstdefinestyle{prompt}{
    basicstyle=\ttfamily\fontsize{7pt}{8pt}\selectfont,
    frame=none,
    breaklines=true,
    backgroundcolor=\color{lightgray},
    breakatwhitespace=true,
    breakindent=0pt,
    escapeinside={(*@}{@*)},
    numbers=none,
    numbersep=5pt,
    xleftmargin=5pt,
    aboveskip=2pt,
    belowskip=2pt,
}
\tcbset{
  aibox/.style={
    top=10pt,
    colback=white,
    center,
  }
}
\newtcolorbox{AIbox}[2][]{aibox, title=#2,#1}

\clearpage
\section*{Appendix}

\section{Dataset Details}
\noindent
\textbf{Encyclopedic VQA} dataset serves as a rigorous benchmark for knowledge-intensive visual reasoning, comprising approximately 1 million VQA samples derived from 221,000 unique question-answer pairs. These samples are centered around 16.7k distinct fine-grained entities, with visual data sourced from the iNaturalist 2021 and Google Landmarks Dataset V2. To facilitate evidence-based answering, the dataset is coupled with a comprehensive, controlled knowledge base consisting of 2 million Wikipedia articles from the WikiWeb2M corpus, each enriched with supporting images to provide necessary context. While the full dataset encompasses both single-hop and complex multi-hop reasoning tasks, this study focuses exclusively on the single-hop subset and utilizes the officially released 2M knowledge base to evaluate the model's capability in retrieving and synthesizing external encyclopedic information for precise visual recognition and fact-based answering.

\noindent
\textbf{InfoSeek} benchmark is a large-scale dataset designed to challenge models in visual information seeking across a diverse spectrum of more than 11,000 visual entities derived from the OVEN framework. To balance the need for extensive training data with the necessity for rigorous evaluation, the dataset is structured into a massive automatically generated training set containing 1.3 million questions and a high-quality, human-annotated evaluation set comprising 8.9 thousand samples. Since the original 6-million-entity knowledge base was not publicly released, we adhere to the experimental protocols established by EchoSight, by utilizing their released 100,000-entity knowledge base filtered from Wikipedia. This ensures a consistent and fair comparison within the research community, specifically focusing on the model's ability to retrieve relevant encyclopedic evidence and generalize to both seen and unseen entities in a real-world information-seeking context.

\noindent
\textbf{M2KR} benchmark is a comprehensive training and evaluation framework designed to foster the development of general-purpose multimodal retrievers by repurposing nine diverse vision-and-language datasets into a consistent, uniform retrieval format. This benchmark suite encompasses three fundamental task types: Image-to-Text (I2T), Question-to-Text (Q2T), and the more complex Image-Question-to-Text (IQ2T), which requires a model to jointly synthesize visual and textual information to accurately retrieve relevant documents. In our experimental framework, we concentrate on these knowledge-intensive IQ2T subtasks, specifically utilizing high-quality datasets such as E-VQA for its focus on fine-grained entity properties and specialized domain knowledge, OKVQA for its requirement of outside world knowledge, and OVEN for its emphasis on open-domain visual entity recognition. By standardizing these datasets with task-specific prompting instructions, M2KR provides a robust environment for evaluating a retriever's ability to navigate large-scale external knowledge bases, such as the WikiWeb2M corpus, ensuring that models can effectively bridge the gap between complex visual understanding and factual knowledge retrieval.

\begin{figure}[htbp]
\centering
\includegraphics[width=1.0\columnwidth]{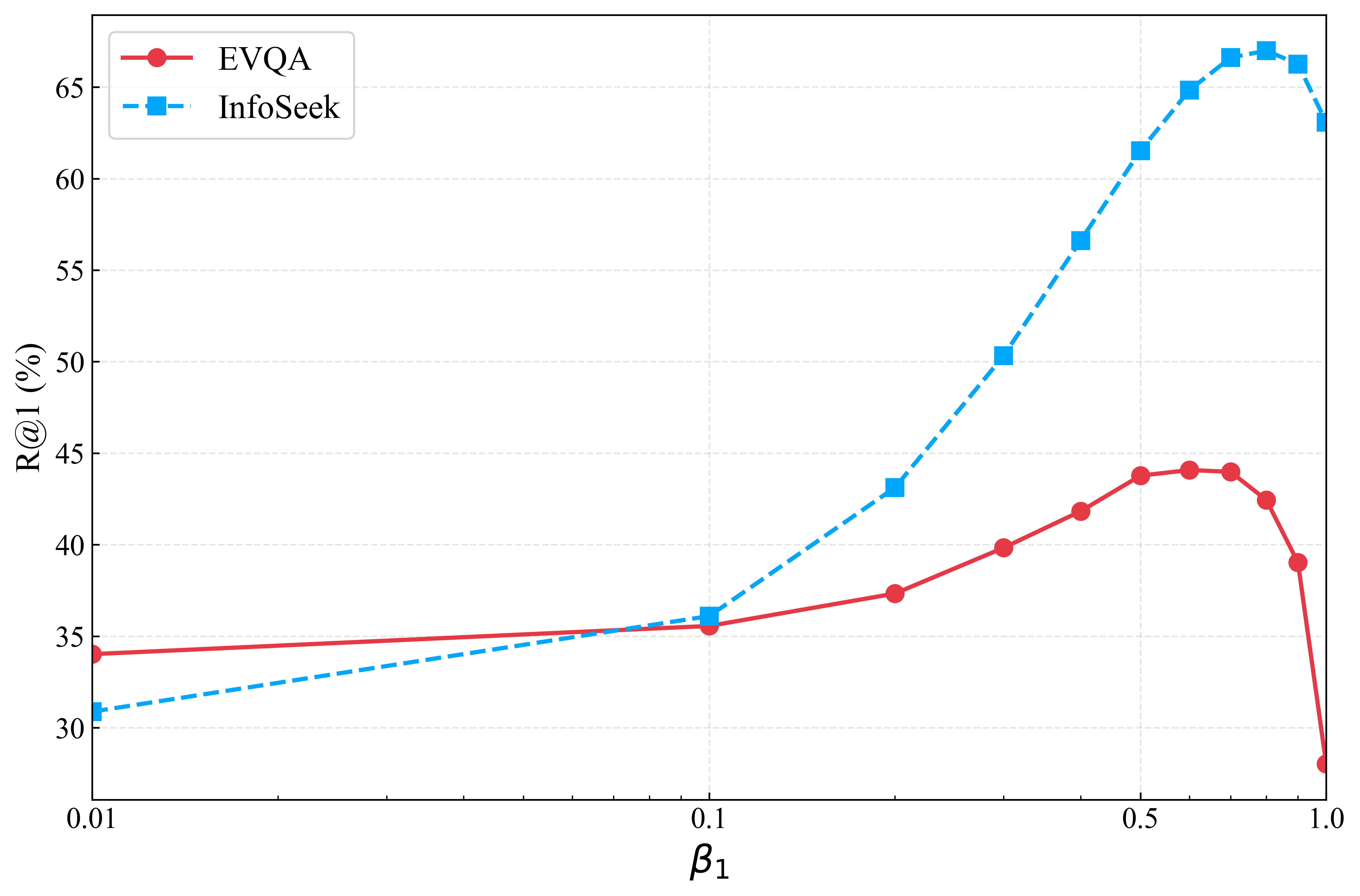}
\caption{Impact of the multi-modal rerank stage weight $\beta_1$ on the final retrieval accuracy (R@1). The x-axis (log scale) represents the weight assigned to the initial retrieval similarity score, where $\beta_1=1.0$ indicates using only the retrieval score without reranking. The results show that a hybrid scoring mechanism yields the best performance, with optimal weights of 0.6 for EVQA and 0.8 for InfoSeek.}
\label{supp:fig2}
\end{figure}

\section{Parameter Searching}
\label{sec:appendix}

\begin{figure*}[htbp]
\centering
\includegraphics[width=0.9\textwidth]{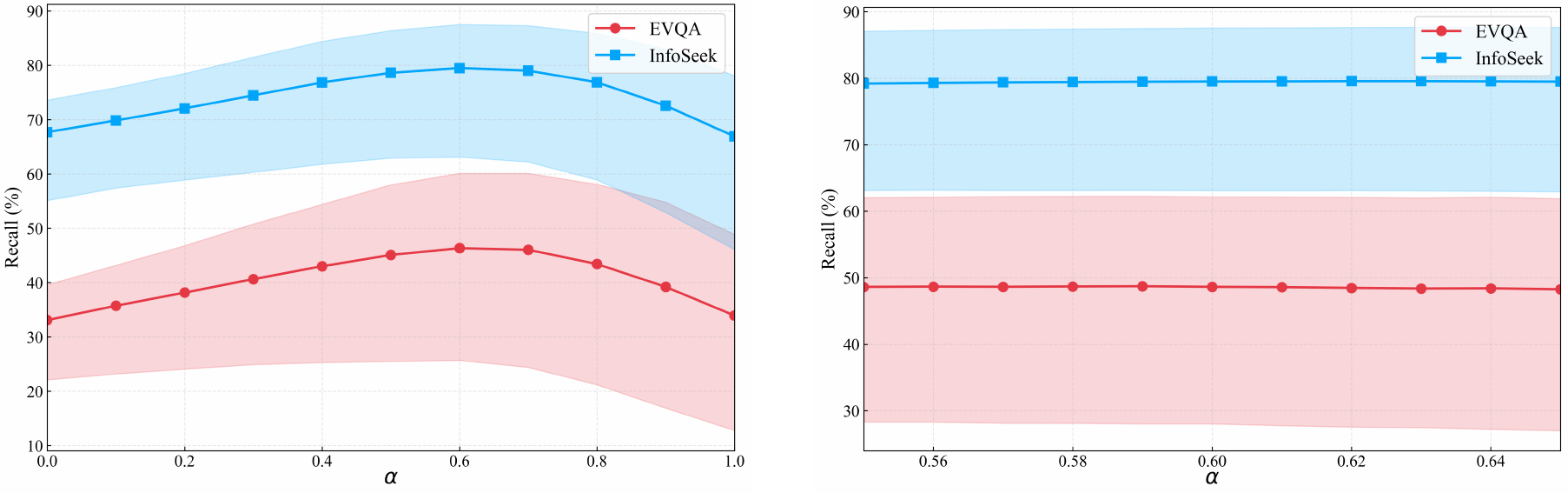}
\caption{Impact of the modality weighting hyperparameter $\alpha$ on retrieval performance using refined queries. Left: Coarse-grained search across the full range $\alpha \in [0, 1]$, demonstrating that multi-modal retrieval outperforms uni-modal baselines. Right: Fine-grained search within the interval $[0.55, 0.65]$ to identify the precise optimal weights. The central markers denote the average recall (mean of R@1, R@5, R@10, and R@20), while the shaded areas indicate the performance range between the lower bound (R@1) and the upper bound (R@20).}
\label{supp:fig1}
\end{figure*}

\subsection{Retrieval Modality Weight}
To determine the optimal fusion ratio between visual and textual modalities, we conduct a two-stage parameter search for the weighting hyperparameter $\alpha$. As depicted in the left panel of Figure~\ref{supp:fig1}, the coarse-grained search across the full range $\alpha \in [0, 1]$ reveals a distinct performance peak around 0.6. This trend confirms that the proposed hybrid retrieval strategy significantly outperforms uni-modal approaches (where $\alpha=0$ or $\alpha=1$). Furthermore, the results indicate a preference for visual semantics, as the performance drops more sharply when $\alpha < 0.5$ compared to $\alpha > 0.5$. To pinpoint the exact optimal configuration, we perform a fine-grained grid search within the interval $[0.55, 0.65]$ with a step size of 0.01, as shown in the right panel of Figure~\ref{supp:fig1}. Based on the peak recall values observed in this refined search space, we set the optimal hyperparameters as $\alpha=0.59$ for EVQA and $\alpha=0.63$ for InfoSeek.

\subsection{Multi-modal Rerank Stage Weight}
To effectively aggregate the global semantic matching capability of the dense retriever and the local discriminative power of the multi-modal reranker, we introduce a hyperparameter $\beta_1$ to balance their respective scores. Specifically, the final ranking score is a weighted sum where $\beta_1$ controls the contribution of the initial retrieval cosine similarity, and $1-\beta_1$ controls the contribution of the reranking score. Notably, setting $\beta_1=1.0$ is equivalent to using the raw retrieval score exclusively (i.e., disabling the reranker).

Figure~\ref{supp:fig2} illustrates the sensitivity of the Top-1 Recall (R@1) to varying $\beta_1$ values on both EVQA and InfoSeek datasets. We observe a consistent trend where performance improves as $\beta_1$ increases from extremely low values (0.01), suggesting that the reranker's score alone is insufficient and requires the global context provided by the retrieval score as an anchor. The performance peaks when a balanced integration is achieved—specifically at $\beta_1=0.6$ for EVQA and $\beta_1=0.8$ for InfoSeek. Beyond these optimal points, particularly as $\beta_1$ approaches 1.0, the performance drops, confirming that the multi-modal reranker provides critical fine-grained filtering that the dense retriever alone cannot achieve.

\subsection{Textual Rerank Stage Weight}
Following the multi-modal reranking phase, we isolate the top-ranked candidate section and retrieve its corresponding full Wikipedia article. To further pinpoint the precise evidence, we apply a textual reranker to score all sections within this article. The final selection of the top-1 section is determined by a weighted fusion of the initial multi-modal reranking score and the intra-article textual reranking score, controlled by the hyperparameter $\beta_2$.

\begin{figure}[htbp]
\centering
\includegraphics[width=1.0\columnwidth]{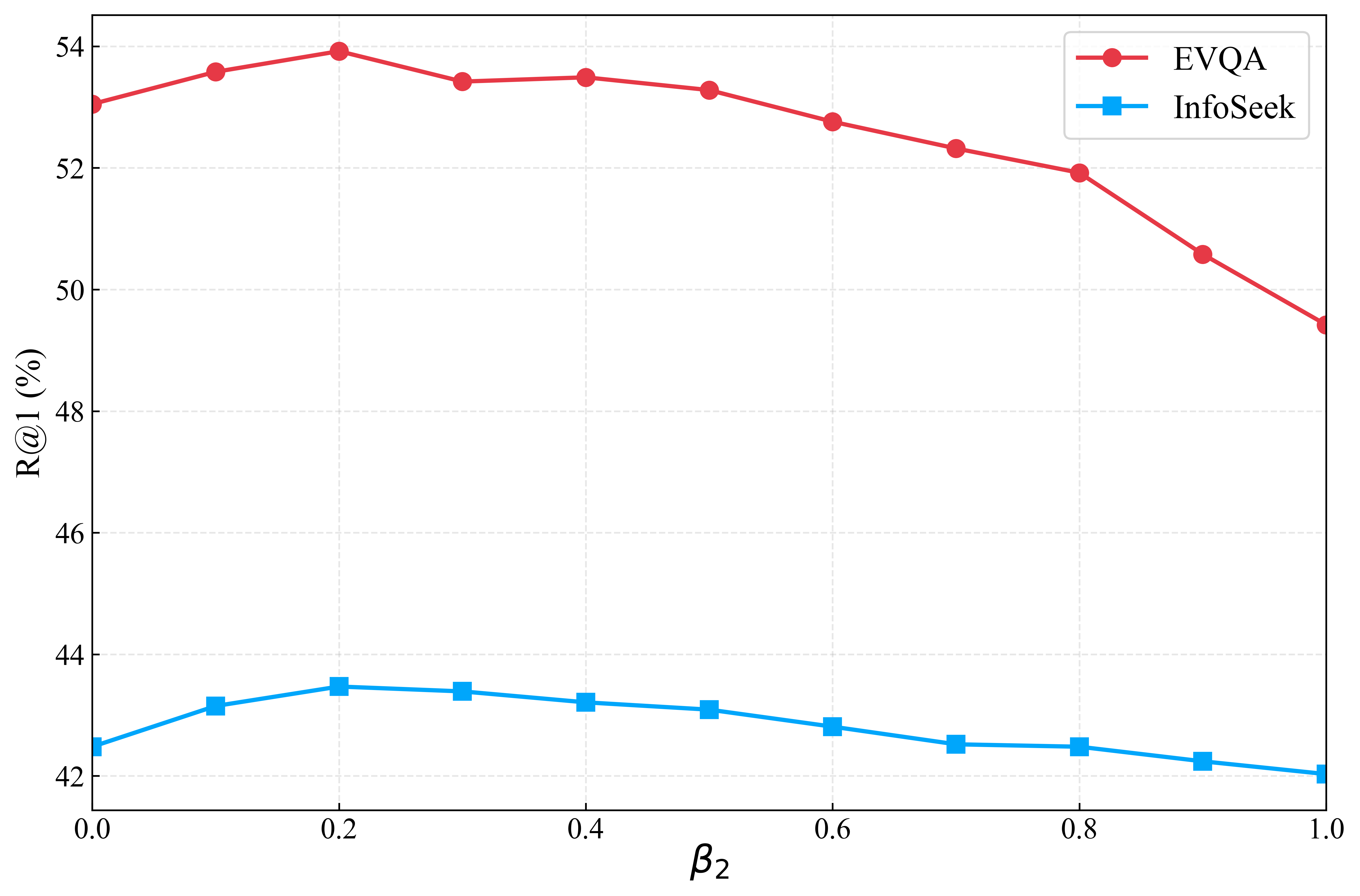}
\caption{Impact of the textual rerank stage weight $\beta_2$ on the final context retrieval accuracy (Recall@1). The results demonstrate that a moderate incorporation of textual reranking scores ($\beta_2=0.2$) yields the optimal performance for both EVQA and InfoSeek datasets.}
\label{supp:fig3}
\end{figure}

Figure~\ref{supp:fig3} illustrates the impact of $\beta_2$ on the final retrieval accuracy (Recall@1). We observe that performance initially improves as $\beta_2$ increases from 0, indicating that incorporating fine-grained textual verification helps filter out noise within the candidate article. However, the performance peaks at $\beta_2=0.2$ for both EVQA (53.92\%) and InfoSeek (43.47\%) and subsequently declines as the weight increases further. This trend suggests that while local textual coherence is a valuable auxiliary signal, the global semantic alignment captured by the multi-modal reranker remains the primary indicator of relevance. Excessive reliance on the textual score dilutes this global context, leading to suboptimal retrieval. Consequently, we set $\beta_2=0.2$ as the default configuration for our framework.

\section{More Experimental Details}

\subsection{More details on M2KR benchmark}
To provide a comprehensive evaluation of WikiSeeker's generalization capabilities, we present fine-grained retrieval results on the EVQA, OKVQA, and OVEN splits of the M2KR benchmark. We extend the evaluation metrics to cover a wider range of retrieval depths (up to $K=100$ or $K=500$) to analyze both top-rank precision and long-tail coverage.

\noindent
\textbf{Performance on EVQA.} As detailed in Table~\ref{supp:tab:m2kr_detailed_evqa}, integrating the Refiner yields consistent performance improvements across all metrics for the EVQA split. The gains are observed in both standard Recall and Pseudo Recall, indicating that the refined queries help identify not only the ground-truth sections but also semantically equivalent candidates that may have been missed by the baseline.

\noindent
\textbf{Performance on OKVQA.} Table~\ref{supp:tab:m2kr_detailed_okvqa} presents the results on the OKVQA split. The improvements here are particularly pronounced. The Refiner boosts Pseudo Recall@1 from 42.01\% to 57.91\% and Recall@1 from 13.77\% to 20.69\%. These substantial margins suggest that the visual-only baseline often fails to capture the necessary commonsense knowledge required by OKVQA, whereas our query refinement successfully bridges this semantic gap, significantly enhancing the relevance of top-ranked results.

\noindent
\textbf{Performance on OVEN.} Finally, we report the performance on the OVEN split in Table~\ref{supp:tab:m2kr_oven}. Since OVEN is an entity-centric open-domain dataset, precise immediate retrieval is critical. Our method achieves a remarkable gain in Recall@1, increasing from 31.09\% to 42.75\%. Furthermore, the performance advantage persists even at extended retrieval depths (Recall@500), demonstrating that WikiSeeker effectively enriches the candidate pool with correct entities even in challenging search scenarios.
\begin{table}[htbp]
\centering
\begin{tabular}{lcc}
\toprule
\textbf{Dataset} & \textbf{PASS Samples} & \textbf{FAIL Samples} \\
\midrule
EVQA & 3,000 & 5,000 \\
InfoSeek & 15,000 & 15,000 \\
\midrule
\textbf{Total} & \textbf{18,000} & \textbf{20,000} \\
\bottomrule
\end{tabular}
\caption{Distribution of the training data used for fine-tuning the Inspector module. "PASS" samples represent successful retrieval cases where the context is sufficient, while "FAIL" samples represent cases where the Inspector must rely on internal parametric knowledge.}
\label{supp:tab:inspector_data}
\end{table}

\begin{table*}[htbp]
\centering
\resizebox{\textwidth}{!}{%
\begin{tabular}{lcccccccccccc}
\toprule
\multirow{2}{*}{\textbf{Method}} & \multicolumn{6}{c}{\textbf{Pseudo Recall@K}} & \multicolumn{6}{c}{\textbf{Recall@K}} \\
\cmidrule(lr){2-7} \cmidrule(lr){8-13}
 & \textbf{1} & \textbf{5} & \textbf{10} & \textbf{20} & \textbf{50} & \textbf{100} & \textbf{1} & \textbf{5} & \textbf{10} & \textbf{20} & \textbf{50} & \textbf{100} \\
\midrule
w/o. Refiner & 51.20 & 72.24 & 78.72 & 83.44 & 88.37 & 90.93 & 40.40 & 62.53 & 70.24 & 78.03 & 86.05 & 91.09 \\
w. Refiner & \textbf{51.57} & \textbf{72.48} & \textbf{79.20} & \textbf{84.32} & \textbf{88.91} & \textbf{91.07} & \textbf{43.12} & \textbf{65.25} & \textbf{73.17} & \textbf{79.95} & \textbf{87.01} & \textbf{91.25} \\
\bottomrule
\end{tabular}%
}
\caption{Detailed retrieval performance comparison on the EVQA split of the M2KR benchmark.}
\label{supp:tab:m2kr_detailed_evqa}
\end{table*}

\begin{table*}[htbp]
\centering
\resizebox{\textwidth}{!}{%
\begin{tabular}{lcccccccccccc}
\toprule
\multirow{2}{*}{\textbf{Method}} & \multicolumn{6}{c}{\textbf{Pseudo Recall@K}} & \multicolumn{6}{c}{\textbf{Recall@K}} \\
\cmidrule(lr){2-7} \cmidrule(lr){8-13}
 & \textbf{1} & \textbf{5} & \textbf{10} & \textbf{20} & \textbf{50} & \textbf{100} & \textbf{1} & \textbf{5} & \textbf{10} & \textbf{20} & \textbf{50} & \textbf{100} \\
\midrule
w/o. Refiner & 42.01 & 67.52 & 78.02 & 86.64 & 94.11 & 96.87 & 13.77 & 30.52 & 39.34 & 49.56 & 63.79 & 71.68 \\
w. Refiner & \textbf{57.91} & \textbf{77.55} & \textbf{85.32} & \textbf{91.58} & \textbf{96.43} & \textbf{98.18} & \textbf{20.69} & \textbf{41.48} & \textbf{51.15} & \textbf{61.57} & \textbf{72.65} & \textbf{79.45} \\
\bottomrule
\end{tabular}%
}
\caption{Detailed retrieval performance comparison on the OKVQA split of the M2KR benchmark.}
\label{supp:tab:m2kr_detailed_okvqa}
\end{table*}

\begin{table}[ht]
\centering
\resizebox{\columnwidth}{!}{%
\begin{tabular}{lccccccc}
\toprule
\multirow{2}{*}{\textbf{Method}} & \multicolumn{7}{c}{\textbf{Recall@K}} \\
\cmidrule(lr){2-8}
 & \textbf{1} & \textbf{5} & \textbf{10} & \textbf{20} & \textbf{50} & \textbf{100} & \textbf{500} \\
\midrule
w/o. Refiner & 31.09 & 64.18 & 76.46 & 85.25 & 92.52 & 95.21 & 97.97 \\
w. Refiner & \textbf{42.75} & \textbf{69.71} & \textbf{78.85} & \textbf{86.89} & \textbf{92.73} & \textbf{95.72} & \textbf{98.34} \\
\bottomrule
\end{tabular}%
}
\caption{Detailed retrieval performance on the OVEN split of the M2KR benchmark.}
\label{supp:tab:m2kr_oven}
\end{table}

\subsection{More details about Limitations of VLMs as Answer Generators}
\label{supp:limitVLM}
In the main text, we discussed the limitations of VLMs as answer generators compared to LLMs when provided with textual context. Table~\ref{supp:tab:snr_full} provides the complete set of results for this experiment across the full spectrum of Signal-to-Noise Ratios (SNR), ranging from 0 (completely irrelevant context) to 1.0 (oracle context). The experimental results are entirely consistent with the analysis presented in the main text.

\begin{table*}[t]
\centering
\resizebox{\textwidth}{!}{%
\begin{tabular}{lccccccccccc}
\toprule
\multirow{2}{*}{\textbf{Method}} & \multicolumn{11}{c}{\textbf{Signal-to-Noise Ratio}} \\
\cmidrule(lr){2-12}
 & \textbf{0} & \textbf{0.1} & \textbf{0.2} & \textbf{0.3} & \textbf{0.4} & \textbf{0.5} & \textbf{0.6} & \textbf{0.7} & \textbf{0.8} & \textbf{0.9} & \textbf{Oracle} \\
\midrule
Qwen2.5-VL-7B (I+T) & \textbf{22.75} & \textbf{28.27} & \textbf{35.45} & 41.26 & 48.27 & 55.10 & 61.68 & 68.99 & 74.99 & 81.79 & 88.46 \\
Qwen2.5-VL-7B (Text-only) & 19.44 & 27.18 & 34.91 & 41.75 & 49.20 & 56.13 & 62.93 & 70.82 & 77.33 & 84.42 & 91.10 \\
LLaMA3-8B & 18.51 & 25.81 & 33.90 & 40.48 & 48.80 & 55.92 & 63.12 & \textbf{70.88} & \textbf{78.08} & \textbf{85.37} & 92.99 \\
Qwen2.5-7B & 19.77 & 26.17 & 34.42 & \textbf{42.52} & \textbf{49.31} & \textbf{56.53} & \textbf{63.37} & 70.80 & 77.77 & 85.18 & \textbf{93.45} \\
\bottomrule
\end{tabular}%
}
\caption{Full VQA performance comparison under different Signal-to-Noise Ratios. The ratio represents the proportion of correct sections mixed into the input context. "Qwen2.5-VL-7B (I+T)" denotes the standard multimodal setting, while "(Text-only)" denotes the same model with the image input removed. The best result for each ratio is highlighted in bold.}
\label{supp:tab:snr_full}
\end{table*}

\subsection{Training Dataset of Inspector}
\label{supp:inspector}
To equip the Inspector with the robust capability to discriminate between sufficient and insufficient contexts, we construct a comprehensive instruction-tuning dataset comprising 38,000 samples derived from the training splits of both EVQA and InfoSeek benchmarks. As outlined in Table~\ref{supp:tab:inspector_data}, this dataset is carefully balanced between positive ("PASS") samples, where the retrieved context contains the necessary evidence, and negative ("FAIL") samples, where the Inspector determines the context is inadequate and must rely on its internal parametric knowledge. Specifically, we aggregate 8,000 samples from EVQA (3,000 PASS and 5,000 FAIL) and 30,000 samples from InfoSeek (15,000 PASS and 15,000 FAIL) to ensure diversity in visual complexity and question types.

The data labeling process employs distinct strategies for each benchmark to accommodate differences in annotation availability. For EVQA, which provides explicit ground-truth section annotations, we label a retrieved context as "PASS" if the top-ranked section matches the ground truth; otherwise, it is labeled as "FAIL," where we employ an auxiliary LLM to expand the short ground-truth answer into a descriptive sentence to serve as the VLM's target response. In contrast, since InfoSeek lacks granular section annotations and only provides ground-truth entities, we implement a rigorous proxy validation strategy. We consider a retrieved context valid ("PASS") only if it successfully recalls the ground-truth entity and simultaneously enables a zero-shot LLM generator to derive the correct answer. Samples failing these dual criteria are categorized as "FAIL," thereby ensuring high-quality supervision for the Inspector's validation mechanism.

\subsection{Efficiency Analysis}
\label{supp:efficiency}
In this section, we discuss the training and inference efficiency of our framework. Regarding training efficiency, our approach is designed to be computationally lean. For the Refiner, we utilize GRPO, which eliminates the need for a value model required by standard PPO algorithms, thereby reducing memory usage and computational overhead during the RL training phase. Furthermore, the fine-tuning of our Generator and Inspector is implemented via the efficient LLaMA-Factory framework, ensuring optimized parameter updates with minimal resource consumption.

To evaluate inference efficiency, we conducted a comparative study on the M2KR EVQA split. We benchmarked WikiSeeker against two baselines: the direct generation method (LLaVA-1.5-7B) and the one-step multimodal RAG method (PreFLMR). We measured three key metrics: Average Retrieval Time, Average Inference Time, and VQA Performance. The results are summarized in Table \ref{tab:efficiency}.

\begin{table}[htbp]
\centering
\resizebox{\columnwidth}{!}{
\begin{tabular}{lccc}
\toprule
\textbf{Method} & \textbf{Avg. Ret. Time} & \textbf{Avg. Inf. Time} & \textbf{VQA Result} \\
\midrule
LLaVA-1.5-7B & - & 1.432 & 17.00 \\
PreFLMR & 0.984 & 2.196 & 54.45 \\
WikiSeeker (Ours) & 0.915 & 2.417 & 65.87 \\
\bottomrule
\end{tabular}
}
\caption{Inference efficiency comparison on the M2KR EVQA split.}
\label{tab:efficiency}
\end{table}

As shown in Table \ref{tab:efficiency}, compared to PreFLMR, our framework demonstrates faster retrieval speeds, even accounting for the additional step of VLM-based query refinement. While the total inference time exhibits a marginal increase due to the multi-agent architecture, WikiSeeker achieves substantial improvements in VQA performance, significantly surpassing both LLaVA-1.5-7B and PreFLMR. These results demonstrate that our framework offers a highly favorable trade-off, achieving SOTA performance while maintaining competitive efficiency.

\subsection{Evaluation of the Inspector Module}
\label{sec:inspector_eval}

To substantiate its reliability, we measure the Inspector's performance as a standalone binary classification process based on Equation \ref{eq:6}. We conducted a comprehensive evaluation on the EVQA test set. The Inspector's performance in determining whether re-ranked sections are relevant (PASS/FAIL) is summarized in the confusion matrix in Table \ref{tab:inspector_cm}.

\begin{table}[htbp]
\centering
\resizebox{\columnwidth}{!}{
\begin{tabular}{lcc}
\toprule
& \textbf{Ground Truth: PASS} & \textbf{Ground Truth: FAIL} \\
\midrule
\textbf{Predicted: PASS} & TP: 1,274 (26.82\%) & FP: 264 (5.60\%) \\
\textbf{Predicted: FAIL} & FN: 586 (12.34\%) & TN: 2,626 (55.28\%) \\
\bottomrule
\end{tabular}
}
\caption{Confusion matrix of the Inspector's routing decisions on the EVQA test set.}
\label{tab:inspector_cm}
\end{table}

The Inspector achieves an overall routing accuracy of 82.1\%, demonstrating its proficiency in validating retrieved context. Notably, the False Negative (FN) rate is higher than the False Positive (FP) rate (12.34\% vs. 5.60\%), reflecting a conservative routing strategy. Comparing the two failure modes, the consequences of an FP result are particularly detrimental, as they feed noisy or irrelevant context to a text-only LLM, directly leading to hallucinations. In contrast, FN scenarios represent a much safer failure mode: the query is simply routed back to the VLM (Inspector). As specified in our Inspector prompt (Figure \ref{fig: inspector_prompt}), the VLM in this path still retains access to both the input image and the retrieved context. Given the VLM's robust internal parametric knowledge and visual reasoning capabilities, it remains highly likely to produce a correct answer.

\section{Prompt Used in WikiSeeker}
\label{supp:prompt}
In this section, we provide a detailed overview of the prompts employed across the various modules of WikiSeeker to facilitate reproducibility. The system prompt and reasoning instructions for the Refiner, designed to expand user queries using visual semantics, are presented in Figure~\ref{fig: refiner_prompt}. The prompt for the Inspector, which is tasked with determining the consistency and sufficiency of the retrieved context, is illustrated in Figure~\ref{fig: inspector_prompt}.

For the LLM Generator, we adopt prompt templates that are largely consistent with those used in the OMGM. The specific prompts tailored for the EVQA and InfoSeek datasets are shown in Figure~\ref{fig: generator_prompt1} and Figure~\ref{fig: generator_prompt2}, respectively. Regarding data processing and knowledge base construction, Figure~\ref{fig: expansion_prompt} displays the prompt used to expand short ground-truth answers into natural sentences for constructing the Inspector's training data. Figure~\ref{fig: summary} depicts the prompt applied to concisely summarize excessively long Wikipedia sections during the construction of our multi-modal knowledge base. Finally, Figure~\ref{fig: caption_prompt} outlines the specific prompts utilized for the image captioning and zero-shot query expansion variants in our ablation studies.

\begin{figure*}[htbp] 
\begin{AIbox}{WikiSeeker Refiner Prompt.}
{\color{black}\bf \large System Prompt:} 
\vspace{1mm}
\\
\textbf{Character Introduction}  \\
You are an expert in generating queries for encyclopedic retrieval. You first think about the reasoning process in the mind and then provide the user with the answer. Given a question about the given image <image>, your task is to retain the original query while expanding it with additional relevant information derived from both the visual content and world knowledge, to retrieve documents that best answer the question.

\textbf{Response Format}  \\
Show your work in <think> </think> tags. Your final response must be in JSON format within <answer> </answer> tags. For example,
\begin{lstlisting}[style=prompt]
<answer>
{
    "query": "...."
} 
</answer>. 
\end{lstlisting}

\tcblower
{\color{black}\bf \large User Prompt:}\\
Here's the user query: {\color{deepblue}\bf \{Query\}}
Assistant: Let me think step by step. <think>

\end{AIbox}
\vspace{-1em}
\caption{Prompt of WikiSeeker Refiner.}
\label{fig: refiner_prompt}
\end{figure*}

\begin{figure*}[htbp] 
\begin{AIbox}{WikiSeeker Inspector Prompt.}
{\color{black}\bf \large System Prompt:} 
\vspace{1mm}
\\
\textbf{Character Introduction}  \\
You are an assistant to determine the consistency and completeness of the provided context in relation to a question and an image <image>. You will receive a question and a retrieved context. Follow these steps:\\
1. Check if the context is consistent with both the image and the question. \\
2. Determine if the context contains the answer to the question. \\
\textbf{Response Format}  \\
If both conditions are satisfied, respond with:
\begin{lstlisting}[style=prompt]
{
    "pass": "true"
} 
\end{lstlisting}
If either condition is not satisfied, respond with:
\begin{lstlisting}[style=prompt]
{
    "pass": "false",
    "answer": "predicted answer"
} 
\end{lstlisting}
Be concise and ensure your responses are in JSON format.

\tcblower
{\color{black}\bf \large User Prompt:}\\
Question: {\color{deepblue}\bf \{Query\}} \\
Retrieved Context:  {\color{deepblue}\bf \{Context\}}

\end{AIbox}
\vspace{-1em}
\caption{Prompt of WikiSeeker Inspector.}
\label{fig: inspector_prompt}
\end{figure*}

\begin{figure*}[htbp] 
\begin{AIbox}{WikiSeeker Generator Prompt on EVQA.}
{\color{black}\bf \large System Prompt:} 
\vspace{1mm}
\\
You are a helpful assistant for answering encyclopedic questions.If the context does not contain the information required to answer the question, you should answer the question using internal model knowledge. 
\tcblower
{\color{black}\bf \large User Prompt:}\\
Context: {\color{deepblue}\bf \{Context\}} \\
Question:  {\color{deepblue}\bf \{Question\}} \\
The answer is:

\end{AIbox}
\vspace{-1em}
\caption{Prompt of WikiSeeker Generator on EVQA.}
\label{fig: generator_prompt1}
\end{figure*}

\begin{figure*}[!ht] 
\begin{AIbox}{WikiSeeker Generator Prompt on InfoSeek.}
{\color{black}\bf \large System Prompt:} 
\vspace{1mm}
\\
You are a helpful assistant for answering encyclopedic questions. Do not answer anything else.If you need to answer questions about numbers or time, please output the corresponding numerical format directly.If the context does not contain the information required to answer the question, you should answer the question using internal model knowledge.
\tcblower
{\color{black}\bf \large User Prompt:}\\
Context: {\color{deepblue}\bf \{Context\}} \\
Question:  {\color{deepblue}\bf \{Question\}} \\
Just answer the questions , no explanations needed. Short answer is:

\end{AIbox}
\vspace{-1em}
\caption{Prompt of WikiSeeker Generator on InfoSeek.}
\label{fig: generator_prompt2}
\end{figure*}

\begin{figure*}[htbp] 
\begin{AIbox}{Prompt for GT answer expansion.}
{\color{black}\bf \large System Prompt:} 
\vspace{1mm}
\\
\textbf{Character Introduction}  \\
Given a question and its short answer, expand the answer into a complete sentence while keeping the original answer intact. Expand the answer into a natural, complete sentence that includes the original answer. 

\textbf{Response Format}  \\
Return your response in JSON format. Output format:
\begin{lstlisting}[style=prompt]
{
    "expanded_answer": "your expanded sentence here"
} 
\end{lstlisting}

\tcblower
{\color{black}\bf \large User Prompt:}\\
Question: {\color{deepblue}\bf \{question\}}
Original Answer: {\color{deepblue}\bf \{original answer\}}

\end{AIbox}
\vspace{-1em}
\caption{Prompt for GT answer expansion.}
\label{fig: expansion_prompt}
\end{figure*}

\begin{figure*}[htbp] 
\begin{AIbox}{Prompt designed for summarizing extended Wikipedia sections.}
{\color{black}\bf \large System Prompt:} 
\vspace{1mm}
\\
Summarize the following Wikipedia section concisely while preserving key information.
\tcblower
{\color{black}\bf \large User Prompt:}\\
Article: {\color{deepblue}\bf \{title\}} \\
Section:  {\color{deepblue}\bf \{section title\}} \\
Content: {\color{deepblue}\bf \{section text\}} \\
Provide a concise summary:

\end{AIbox}
\vspace{-1em}
\caption{Prompt designed for summarizing extended Wikipedia sections.}
\label{fig: summary}
\end{figure*}

\begin{figure*}[htbp] 
\begin{AIbox}{Prompt for image caption and zero-shot query expansion.}
{\color{black}\bf \large System Prompt:} 
\vspace{1mm}
\\
\textbf{Character Introduction}  \\
You are an expert in generating queries for encyclopedic retrieval. Given a question about the given image, you should: \\
1. Concisely caption the image which is most relevant to the question. \\
2. Retain the original query while expanding it with additional relevant information derived from both the visual content and world knowledge, to retrieve documents that best answer the question. \\
\textbf{Response Format}  \\
Your final response must be in JSON format. For example:
\begin{lstlisting}[style=prompt]
{
    "caption": "...",
    "query": "..." 
} 
\end{lstlisting}

\tcblower
{\color{black}\bf \large User Prompt:}\\
Here's the user query: {\color{deepblue}\bf \{query\}}

\end{AIbox}
\vspace{-1em}
\caption{Prompt for image caption and zero-shot query expansion.}
\label{fig: caption_prompt}
\end{figure*}

\section{Case Study}
To intuitively demonstrate the mechanisms behind WikiSeeker's performance improvements, we provide qualitative visualizations of our two core modules: the Refiner and the Inspector.

\begin{figure*}[htbp]
\centering
\includegraphics[width=1.0\textwidth]{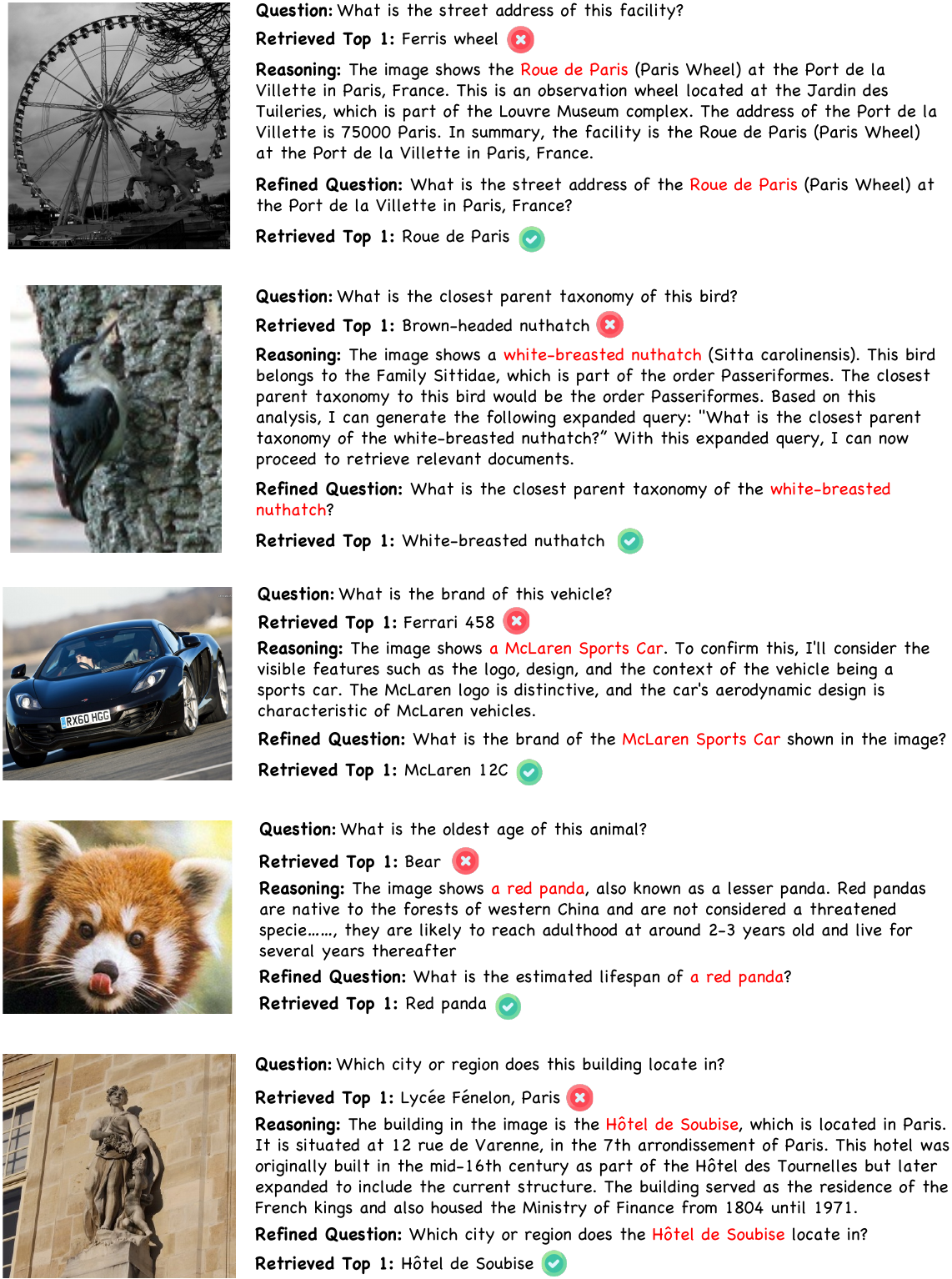}
\caption{Qualitative examples demonstrating the effectiveness of the Refiner. For each case, the original user query fails to retrieve the correct entity (marked with \textcolor{red}{\ding{55}}). The Refiner analyzes the image through a chain-of-thought reasoning process to generate a refined query, which successfully retrieves the ground-truth entity (marked with \textcolor{green}{\ding{51}}).}
\label{supp:fig4}
\end{figure*}

\noindent \textbf{Analysis of the Refiner.} Figure~\ref{supp:fig4} presents five representative examples illustrating the Refiner's workflow. In these scenarios, the original user queries are often too vague or implicit (e.g., "What is the street address of this facility?") to retrieve the correct documents directly, resulting in irrelevant top-1 retrieval results. However, through an explicit reasoning process, the Refiner effectively identifies key visual entities—such as the "Roue de Paris" or the "white-breasted nuthatch"—and incorporates them into an expanded search query. This query expansion allows the Vision-Language Model to actively participate in the retrieval phase, thereby making it significantly easier to retrieve the correct entity.

\begin{figure*}[htbp]
\centering
\includegraphics[width=1.0\textwidth]{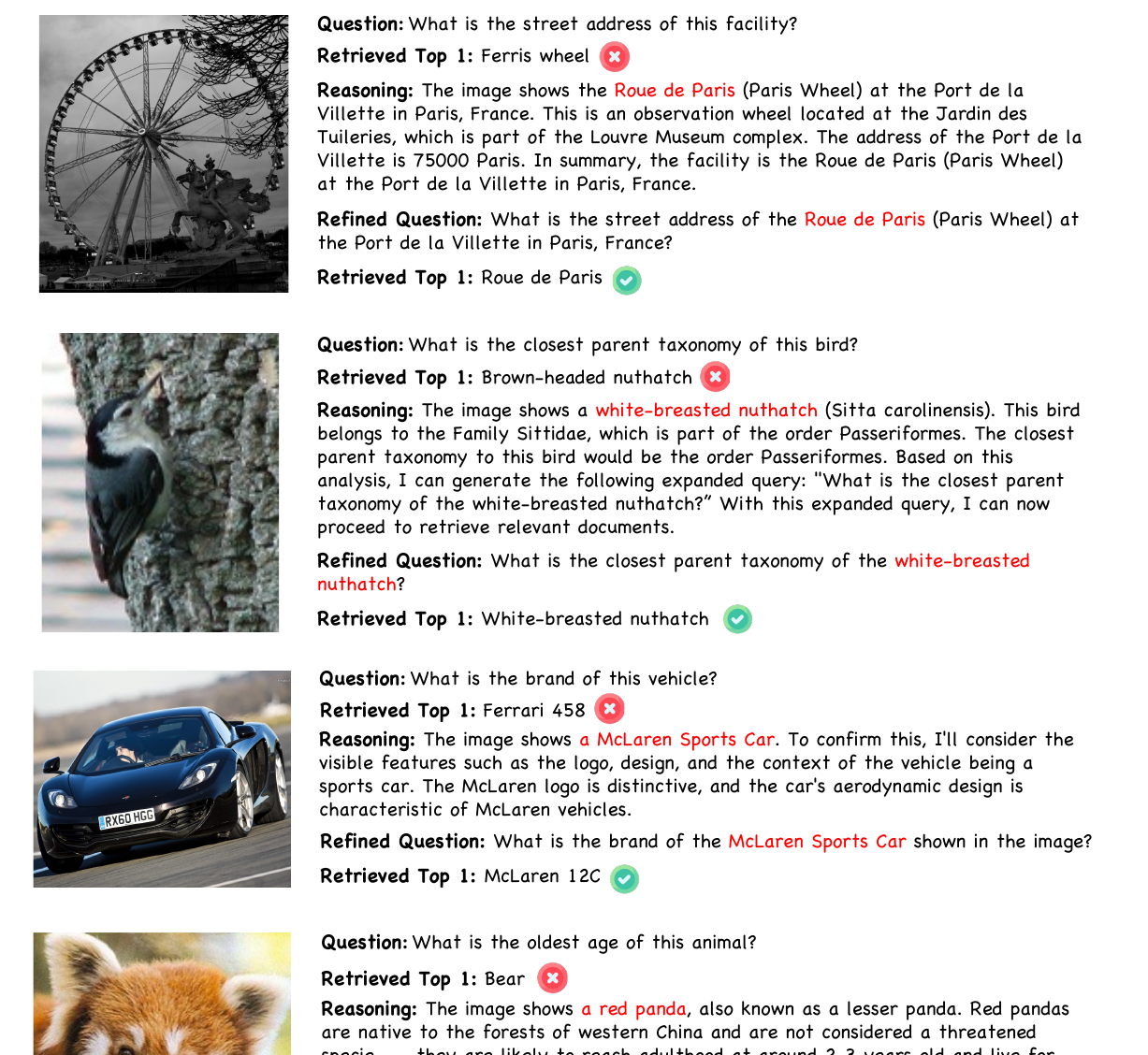}
\caption{Visualization of the decoupled generation strategy enabled by the Inspector. \textbf{Left (Green Context):} Scenarios where retrieval is accurate. The relevant evidence is underlined. The LLM accurately extracts the answer, whereas the VLM fails. The Inspector validates the context and routes the query to the LLM Generator (indicated by ``Gen''). \textbf{Right (Red Context):} Scenarios where retrieval fails. The LLM cannot answer correctly due to missing information, but the VLM correctly answers using internal parametric knowledge. The Inspector detects the retrieval failure and provides the answer directly (indicated by ``Ins'').}
\label{supp:fig5}
\end{figure*}

\noindent
\textbf{Analysis of the Decoupled Generation Strategy.} Figure~\ref{supp:fig5} validates the effectiveness of our decoupled generation strategy and the Inspector's routing logic. We compare the responses of a text-only LLM (Qwen2.5-7B-Instruct) and a VLM (Qwen3-VL-8B-Instruct) against our WikiSeeker framework. 
\begin{itemize} 
\item The \textbf{left column} displays cases where the retrieval is successful (contexts marked in green). Here, the correct answer is explicitly contained within the text (underlined). In these instances, the LLM demonstrates superior reading comprehension, correctly extracting the answer, whereas the VLM fails to focus on the textual evidence. Our Inspector correctly validates these contexts (PASS) and routes the query to the LLM Generator (marked as Gen''), ensuring accuracy. 
\item The \textbf{right column} illustrates retrieval failures (contexts marked in red). Without relevant external knowledge, the LLM hallucinates or fails to answer. In contrast, the VLM successfully answers these questions by leveraging its internal parametric knowledge and visual understanding. Crucially, our Inspector detects the insufficiency of the retrieved context (FAIL) and intervenes by providing the answer directly (marked as Ins''), effectively preventing the propagation of retrieval errors. 
\end{itemize}

\end{document}